\documentclass{article}
\usepackage{iclr2026_conference}
\usepackage{amsmath,amssymb,amsthm}
\usepackage{graphicx}
\usepackage{hyperref}
\usepackage{booktabs}
\usepackage{amssymb}
\newtheorem{theorem}{Theorem}
\newtheorem{lemma}{Lemma}
\newtheorem{proposition}{Proposition}

\title{Murphy’s Laws of AI Alignment: Why the Gap Always Wins}

\author{
\begin{minipage}{0.9\linewidth}
\centering
                Madhava Gaikwad \\
       \texttt{mgaikwad@microsoft.com} \\
                    Microsoft \\
\thanks{This is a working paper open to collab, revision and critique. This work was done in individual capacity and does not represent the views of my employer.}
\end{minipage}
}

\iclrfinalcopy 

\begin{document}
\maketitle

\begin{abstract}
We study reinforcement learning from human feedback under misspecification. Contexts are drawn i.i.d. from a distribution \(\mu\) over \(\mathcal{X}\). There exists a measurable subset \(X_{\mathrm{hard}}\subset\mathcal{X}\) with mass \(\alpha=\mu(X_{\mathrm{hard}})\) on which the feedback channel is systematically biased. We consider pairwise preference feedback and allow adaptive query selection with a budget of \(n\) queries.

We construct two environments \(w\in\{+,-\}\) with reward functions \(r_w\) that differ only on \(X_{\mathrm{hard}}\). Let \(\pi^w\) denote an optimal policy for \(r_w\), and define the separation parameter
\[
\gamma \;=\; \mathbb{E}_{x\sim\mu}\!\left[\big|r_{+}(x,\pi^{+}(x)) - r_{+}(x,\pi^{-}(x))\big|\cdot \mathbf{1}\{x\in X_{\mathrm{hard}}\}\right] \;>\; 0.
\]
On \(X_{\mathrm{hard}}\) the pairwise label favors the truly better action with probability \(\tfrac{1}{2}+w\varepsilon\) for some \(\varepsilon\in(0,\tfrac{1}{2})\). Outside \(X_{\mathrm{hard}}\) feedback is symmetric. For any learner using at most \(n\) queries, Le Cam’s two point method combined with a transcript level Kullback–Leibler decomposition yields
\[
\inf_{\widehat{\pi}}\;\sup_{w\in\{+,-\}} \Big(V_w(\pi^w) - \mathbb{E}_w[V_w(\widehat{\pi})]\Big)
\;\ge\; \frac{\gamma}{4}\exp\!\big(-n\,\alpha\,\kappa(\varepsilon)\big),
\qquad
\kappa(\varepsilon)=4\varepsilon\,\mathrm{atanh}(2\varepsilon).
\]
With access to a calibration oracle \(h(x)=\mathbf{1}\{x\in X_{\mathrm{hard}}\}\) that flags misspecified contexts, an adaptive procedure that concentrates queries on \(X_{\mathrm{hard}}\) achieves expected gap at most \(\eta\) with
\[
Q \;\le\; \frac{1}{2\alpha\varepsilon^{2}}\log\!\frac{\gamma}{\eta}
\]
queries, using a simple majority test on the flagged contexts. The constants arise from per hit Bernoulli KL and standard testing bounds. The results quantify how \(\alpha\) and \(\varepsilon\) govern the sample complexity required to resolve misspecification.
\end{abstract}
\section{Introduction}\label{sec:intro}
Reinforcement learning from human feedback (RLHF) is often instantiated via pairwise comparisons or scalar ratings that train a reward model used for policy optimization. In many deployments the feedback channel is imperfect and the number of informative queries is limited. This paper analyzes the fundamental limitations imposed by misspecification under bounded query budgets and identifies minimal additional structure that removes these limitations.

\paragraph{Setting.}
Let \((\mathcal{X},\mu)\) be a context space with i.i.d. draws \(x\sim\mu\) and a finite action set \(\mathcal{A}\). A policy \(\pi:\mathcal{X}\to\mathcal{A}\) has value \(V_w(\pi)=\mathbb{E}_{x\sim\mu}[r_w(x,\pi(x))]\) under world \(w\). There is a measurable subset \(X_{\mathrm{hard}}\subset\mathcal{X}\) with \(\alpha=\mu(X_{\mathrm{hard}})\) on which feedback is misspecified. Two worlds \(w\in\{+,-\}\) are defined by reward functions \(r_w\) that differ only on \(X_{\mathrm{hard}}\). Let \(\pi^w\) denote an optimal policy for \(r_w\). Define the separation parameter
\[
\gamma \;=\; \mathbb{E}_{x\sim\mu}\!\left[\big|r_{+}(x,\pi^{+}(x)) - r_{+}(x,\pi^{-}(x))\big|\cdot \mathbf{1}\{x\in X_{\mathrm{hard}}\}\right] \;>\; 0.
\]
On \(x\in X_{\mathrm{hard}}\) the pairwise label equals the indicator that the truly better action is preferred, flipped with Massart style bias of magnitude \(\varepsilon\in(0,\tfrac{1}{2})\):
\[
\Pr_{w}\{\text{label favors the truly better action}\mid x\in X_{\mathrm{hard}}\} \;=\; \tfrac{1}{2}+w\varepsilon.
\]
On \(x\in X_{\mathrm{easy}}=\mathcal{X}\setminus X_{\mathrm{hard}}\) the channel is symmetric with probability \(\tfrac{1}{2}\). The learner may adaptively select which queries to issue, up to a budget of \(n\) queries.

\paragraph{Lower bound.}
Let \(\mathsf{P}^{w}\) be the joint law of the full transcript under world \(w\). The transcript level Kullback–Leibler divergence satisfies
\[
D_{\mathrm{KL}}(\mathsf{P}^{+}\,\|\,\mathsf{P}^{-})
\;=\;\sum_{t=1}^{n}\mathbb{E}_{\mathsf{P}^{+}}\!\left[
D_{\mathrm{KL}}\!\big(\mathsf{P}^{+}(Y_t\mid \mathcal{H}_{t-1})\,\|\,\mathsf{P}^{-}(Y_t\mid \mathcal{H}_{t-1})\big)
\right]
\;\le\; n\,\alpha\,\kappa(\varepsilon),
\]
where each term is nonzero only when \(x_t\in X_{\mathrm{hard}}\) and
\[
\kappa(\varepsilon)\;=\;D_{\mathrm{KL}}\!\left(\mathrm{Ber}\!\left(\tfrac{1}{2}+\varepsilon\right)\,\middle\|\,\mathrm{Ber}\!\left(\tfrac{1}{2}-\varepsilon\right)\right)
\;=\; 4\varepsilon\,\mathrm{atanh}(2\varepsilon).
\]
A standard testing inequality converts the KL budget into a lower bound on the Bayes error for distinguishing the two worlds, which implies the finite sample value gap
\[
\inf_{\widehat{\pi}}\;\sup_{w\in\{+,-\}} \Big(V_w(\pi^w) - \mathbb{E}_w[V_w(\widehat{\pi})]\Big)
\;\ge\; \frac{\gamma}{4}\exp\!\big(-n\,\alpha\,\kappa(\varepsilon)\big).
\]
The bound holds for adaptive procedures since the chain rule is applied conditionally on the observed history.

\paragraph{Upper bound with a calibration oracle.}
Suppose an oracle \(h:\mathcal{X}\to\{0,1\}\) indicates membership in \(X_{\mathrm{hard}}\). Draw contexts until \(h(x)=1\). The expected number of draws per hit is \(1/\alpha\). On each hit, collect a bounded preference bit. A majority test over \(T\) hits on \(X_{\mathrm{hard}}\) errs with probability at most \(\exp(-2T\varepsilon^{2})\) by a bounded difference inequality. Setting \(T=\tfrac{1}{2\varepsilon^{2}}\log(\gamma/\eta)\) yields an expected value gap at most \(\eta\) with
\[
Q \;=\; \frac{T}{\alpha} \;\le\; \frac{1}{2\alpha\varepsilon^{2}}\log\!\frac{\gamma}{\eta}
\]
total queries. This matches the dependence on \(\alpha\) and \(\varepsilon\) in the lower bound up to constants.

\paragraph{Discussion.}
The analysis isolates three parameters that govern difficulty under misspecification: prevalence \(\alpha\), bias magnitude \(\varepsilon\), and separation \(\gamma\). The lower bound follows from indistinguishability of two worlds on \(X_{\mathrm{hard}}\), and the upper bound shows that the ability to route queries to that set is sufficient to overcome the barrier. The same blueprint extends to scalar ratings when the per hit log likelihood ratio is bounded by a quantity of order \(\varepsilon^{2}\).

\section{Formal Setup and Main Results}\label{sec:setup}
We study a contextual decision problem with a possibly misspecified preference channel. Contexts are drawn i.i.d. from a probability space \((\mathcal{X},\mathcal{F},\mu)\). The action set \(\mathcal{A}\) is finite. For world \(w\in\{+,-\}\), a reward function \(r_w:\mathcal{X}\times\mathcal{A}\to[0,1]\) induces the value of a policy \(\pi:\mathcal{X}\to\mathcal{A}\),
\[
V_w(\pi)\;=\;\mathbb{E}_{x\sim\mu}\big[r_w(x,\pi(x))\big].
\]

\subsection{Environment family}\label{sec:env}
There exists a measurable subset \(X_{\mathrm{hard}}\subset\mathcal{X}\) with mass \(\alpha:=\mu(X_{\mathrm{hard}})\in(0,1)\), and \(X_{\mathrm{easy}}=\mathcal{X}\setminus X_{\mathrm{hard}}\). The two worlds share the same reward on \(X_{\mathrm{easy}}\) and differ only on \(X_{\mathrm{hard}}\). Let \(\pi^{w}\) be an optimal policy for \(r_w\). Define the separation
\[
\gamma \;=\; \mathbb{E}_{x\sim\mu}\!\left[\big|r_{+}(x,\pi^{+}(x)) - r_{+}(x,\pi^{-}(x))\big|\cdot \mathbf{1}\{x\in X_{\mathrm{hard}}\}\right] \;>\; 0.
\]

\subsection{Feedback channel}\label{sec:channel}
A round \(t\) presents a context \(x_t\sim\mu\). If a pairwise query is issued, a binary label \(Y_t\in\{0,1\}\) is returned. On \(x\in X_{\mathrm{easy}}\), the label is symmetric with mean \(1/2\). On \(x\in X_{\mathrm{hard}}\), the label has Massart type bias of fixed magnitude \(\varepsilon\in(0,\tfrac{1}{2})\):
\[
\Pr_{w}\{\text{label favors the truly better action}\mid x\in X_{\mathrm{hard}}\}\;=\;\tfrac{1}{2}+w\varepsilon.
\]
Scalar ratings admit the same analysis when the per query log likelihood ratio on \(X_{\mathrm{hard}}\) is bounded by a quantity of order \(\varepsilon^{2}\).

\subsection{Learning protocol}\label{sec:protocol}
The learner interacts for at most \(n\) queries. At each \(t\), after observing \(x_t\) and history \(\mathcal{H}_{t-1}\), it decides whether to issue a query. Let \(I_t\in\{0,1\}\) indicate that a query is issued and write the decision as a measurable function \(q_t:\mathcal{X}\times\mathcal{H}_{t-1}\to[0,1]\) with \(\Pr(I_t=1\mid x_t,\mathcal{H}_{t-1})=q_t(x_t,\mathcal{H}_{t-1})\). The transcript law under world \(w\) is \(\mathsf{P}^{w}\). The learner outputs a policy \(\widehat{\pi}\) adapted to the transcript. We evaluate the minimax gap
\[
\Delta_n \;=\; \inf_{\text{alg}}\;\sup_{w\in\{+,-\}}\Big(V_w(\pi^{w})-\mathbb{E}_w[V_w(\widehat{\pi})]\Big).
\]

\subsection{Per hit information}\label{sec:perhit}
\begin{lemma}[Bernoulli KL on opposite biases]\label{lem:bernoulli-kl}
For \(|\varepsilon|<\tfrac{1}{2}\),
\[
\kappa(\varepsilon)\;:=\;D_{\mathrm{KL}}\!\left(\mathrm{Ber}\!\left(\tfrac{1}{2}+\varepsilon\right)\,\middle\|\,\mathrm{Ber}\!\left(\tfrac{1}{2}-\varepsilon\right)\right)
\;=\;4\varepsilon\,\mathrm{atanh}(2\varepsilon)
\;\le\;\frac{8\varepsilon^{2}}{1-4\varepsilon^{2}}.
\]
\emph{Proof.} Write \(p=\tfrac{1}{2}+\varepsilon\) and \(q=\tfrac{1}{2}-\varepsilon\). Then
\[
D_{\mathrm{KL}}(\mathrm{Ber}(p)\|\mathrm{Ber}(q))
= p\log\frac{p}{q}+(1-p)\log\frac{1-p}{1-q}
= (2p-1)\log\frac{1+2\varepsilon}{1-2\varepsilon}
= 2\varepsilon\log\frac{1+2\varepsilon}{1-2\varepsilon}.
\]
Use \(\mathrm{atanh}(z)=\tfrac{1}{2}\log\frac{1+z}{1-z}\) and \(\mathrm{atanh}(z)\le z/(1-z^{2})\) for \(|z|<1\). \(\square\)
\end{lemma}

\subsection{Hit count bound}\label{sec:hitbound}
\begin{lemma}[Expected queried hits on \(X_{\mathrm{hard}}\)]\label{lem:hits}
For any predictable query policy with at most \(n\) queries,
\[
\sum_{t=1}^{n}\mathbb{E}_{\mathsf{P}^{+}}\big[I_t\mathbf{1}\{x_t\in X_{\mathrm{hard}}\}\big]\;\le\; n\,\alpha.
\]
The same bound holds under \(\mathsf{P}^{-}\).
\emph{Proof.} Condition on \(\mathcal{H}_{t-1}\). Since \(x_t\sim\mu\) is independent of \(\mathcal{H}_{t-1}\),
\[
\mathbb{E}[I_t\mathbf{1}\{x_t\in X_{\mathrm{hard}}\}\mid \mathcal{H}_{t-1}]
= \int q_t(x,\mathcal{H}_{t-1})\mathbf{1}\{x\in X_{\mathrm{hard}}\}\,\mu(\mathrm{d}x)
\le \alpha\cdot \sup_{x} q_t(x,\mathcal{H}_{t-1}) \le \alpha.
\]
Summing over \(t\) and taking expectation gives the claim. \(\square\)
\end{lemma}

\subsection{Lower bound}\label{sec:lower}
\begin{theorem}[Finite sample impossibility]\label{thm:lower}
For any adaptive learner using at most \(n\) queries,
\[
\Delta_n \;\ge\; \frac{\gamma}{4}\,\exp\!\big(-n\,\alpha\,\kappa(\varepsilon)\big).
\]
\emph{Proof.} Apply the KL chain rule to the transcript:
\[
D_{\mathrm{KL}}(\mathsf{P}^{+}\,\|\,\mathsf{P}^{-})
=\sum_{t=1}^{n}\mathbb{E}_{\mathsf{P}^{+}}\!\left[
D_{\mathrm{KL}}\big(\mathsf{P}^{+}(Y_t\mid \mathcal{H}_{t-1},x_t,I_t)\,\big\|\,\mathsf{P}^{-}(Y_t\mid \mathcal{H}_{t-1},x_t,I_t)\big)
\right].
\]
The inner KL is zero unless \(I_t=1\) and \(x_t\in X_{\mathrm{hard}}\). In that case it is at most \(\kappa(\varepsilon)\) by Lemma \ref{lem:bernoulli-kl}. Taking expectation and using Lemma \ref{lem:hits} yields
\(D_{\mathrm{KL}}(\mathsf{P}^{+}\,\|\,\mathsf{P}^{-})\le n\,\alpha\,\kappa(\varepsilon)\).
A standard testing inequality gives that the Bayes error for distinguishing the two worlds under equal priors is at least \(\tfrac{1}{4}\exp(-D_{\mathrm{KL}})\). A testing mistake induces value gap at least \(\gamma\). Combine the two relations. \(\square\)
\end{theorem}

\subsection{Oracle upper bound}\label{sec:upper}
\begin{theorem}[Calibration oracle suffices]\label{thm:upper}
Assume access to \(h(x)=\mathbf{1}\{x\in X_{\mathrm{hard}}\}\). For any \(\eta\in(0,\gamma)\) there exists a procedure that uses
\[
Q \;\le\; \frac{1}{2\,\alpha\,\varepsilon^{2}}\log\!\frac{\gamma}{\eta}
\]
queries and returns \(\widehat{\pi}\) such that
\(\sup_{w\in\{+,-\}}\big(V_w(\pi^{w})-\mathbb{E}_w[V_w(\widehat{\pi})]\big)\le\eta\).
\emph{Proof.} Draw contexts until \(h(x)=1\). The expected number of draws per hit is \(1/\alpha\). On each hit issue a pairwise query and record the binary outcome \(Z_i\) with mean \(\tfrac{1}{2}\pm\varepsilon\) depending on \(w\). After \(T\) hits, decide \(\widehat{w}\) by the sign of \(\sum_{i=1}^{T}(Z_i-\tfrac{1}{2})\) and output \(\pi^{\widehat{w}}\). A bounded difference inequality gives
\(\Pr(\widehat{w}\neq w)\le \exp(-2T\varepsilon^{2})\).
Setting \(T=\tfrac{1}{2\varepsilon^{2}}\log(\gamma/\eta)\) makes the expected value gap at most \(\eta\). The expected total queries equal \(Q=T/\alpha\). \(\square\)
\end{theorem}

\subsection{Noisy oracle and minimality}\label{sec:minimal}
\begin{theorem}[Minimality within \(x\)-only binary oracles]\label{thm:minimal}
Let \(\tilde h(x)\in\{0,1\}\) satisfy \(\Pr(\tilde h=1\mid x\in X_{\mathrm{hard}})=\tau\) and \(\Pr(\tilde h=1\mid x\in X_{\mathrm{easy}})=\phi\). Let the procedure keep only \(\tilde h(x)=1\) contexts and run the same majority test on hits. To obtain \(T\) true hits in expectation one needs at least \(T/(\alpha\tau)\) kept draws, hence
\[
Q \;\ge\; \frac{1}{2\,\alpha\,\tau\,\varepsilon^{2}}\log\!\frac{\gamma}{\eta}.
\]
Therefore \(h(x)=\mathbf{1}\{x\in X_{\mathrm{hard}}\}\) with \(\tau=1\) and \(\phi=0\) is minimal in this class.
\emph{Proof.} The fraction of kept draws that are true hits equals
\(\frac{\alpha\tau}{\alpha\tau+(1-\alpha)\phi}\le \tau\). The stated lower bound follows. \(\square\)
\end{theorem}

\subsection{Remark on scalar ratings}\label{sec:ratings}
If on \(X_{\mathrm{hard}}\) the scalar rating \(R\in[0,1]\) satisfies
\(\log\frac{\mathrm{d}\mathsf{P}^{+}}{\mathrm{d}\mathsf{P}^{-}}(R)\) with variance proxy bounded by \(c\varepsilon^{2}\) uniformly in the history, then the per hit KL is \(\le C\varepsilon^{2}\) for a constant \(C\). The proofs above apply with \(\kappa(\varepsilon)\) replaced by \(C\varepsilon^{2}\).
\section{Empirical Indications}\label{sec:empirics}
This section specifies synthetic protocols that reflect the parameters \(\alpha\), \(\varepsilon\), and \(\gamma\), and diagnostics that probe the mechanisms in Theorems \ref{thm:lower} and \ref{thm:upper}. The goal is to visualize the gap predicted by the bound and the effect of concentrating queries on \(X_{\mathrm{hard}}\).

\subsection{Synthetic environment}\label{sec:synth}
Fix dimension \(d\). Draw contexts \(x\in\mathbb{R}^{d}\) i.i.d. from a mixture \(\mu=\,(1-\alpha)\,\mathcal{N}(0,I_d)\,+\,\alpha\,\mathcal{N}(\mu_{\mathrm{h}},I_d)\). Define \(X_{\mathrm{hard}}\) as the support of the second component, so \(\mu(X_{\mathrm{hard}})=\alpha\). Let \(\mathcal{A}=\{a_0,a_1\}\). Choose unit vectors \(\theta,\upsilon\in\mathbb{R}^{d}\) with \(\theta\perp\upsilon\). Set
\[
r_{+}(x,a_1)-r_{+}(x,a_0) \;=\; \mathrm{sign}(\theta^{\top}x)\cdot \mathbf{1}\{x\in X_{\mathrm{easy}}\} \;+\; \mathrm{sign}(\upsilon^{\top}x)\cdot \mathbf{1}\{x\in X_{\mathrm{hard}}\}.
\]
Define \(r_{-}\) by flipping the sign on \(X_{\mathrm{hard}}\). This induces \(\pi^{+}(x)=a_1\) if \(r_{+}(x,a_1)\ge r_{+}(x,a_0)\) and \(\pi^{-}\) analogously. The separation \(\gamma\) is controlled by the margin distribution on \(X_{\mathrm{hard}}\). Pairwise feedback is generated by the Massart model with bias \(\varepsilon\) on \(X_{\mathrm{hard}}\) and symmetry on \(X_{\mathrm{easy}}\).

\subsection{Preference learner and proxy}\label{sec:proxy}
Collect \(n\) pairwise comparisons with adaptive selection. Fit a logistic preference model \(\widehat{s}(x)\) that predicts the advantage of \(a_1\) over \(a_0\). Define the proxy reward \(\widehat{r}(x,a_1)-\widehat{r}(x,a_0)=\widehat{s}(x)\). For a temperature parameter \(\lambda\ge 0\), define a stochastic policy
\[
\pi_{\lambda}(x)=
\begin{cases}
a_1 & \text{with prob. } \sigma\big(\lambda\,\widehat{s}(x)\big),\\
a_0 & \text{otherwise},
\end{cases}
\quad \sigma(u)=(1+e^{-u})^{-1}.
\]
Let \(V_{w}(\pi_{\lambda})\) be the true value under world \(w\) and \(\widehat{V}(\pi_{\lambda})=\mathbb{E}_{x\sim\mu}[\widehat{r}(x,\pi_{\lambda}(x))]\) the proxy value.

\subsection{Diagnostics}\label{sec:diagnostics}
\paragraph{D1. Gap versus optimization pressure.}
Compute \(G(\lambda)=V_{+}(\pi^{+})-V_{+}(\pi_{\lambda})\) and the proxy \(\widehat{G}(\lambda)=\widehat{V}(\pi^{+})-\widehat{V}(\pi_{\lambda})\) for \(\lambda\) on a grid. The theory predicts \(\widehat{V}(\pi_{\lambda})\) increases with \(\lambda\) while \(V_{+}(\pi_{\lambda})\) can plateau or decrease once the policy mass shifts toward \(X_{\mathrm{hard}}\).

\paragraph{D2. In distribution to shifted distribution.}
Construct a shifted test distribution \(\mu_{\rho}\) by increasing the hard mass to \(\alpha_{\rho}=(1+\rho)\alpha\) and renormalizing. Measure \(G_{\rho}(\lambda)\) under \(\mu_{\rho}\). The gap should increase with \(\rho\), consistent with the dependence on \(\alpha\) in Theorem \ref{thm:lower}.

\paragraph{D3. Query routing.}
Implement a calibrated flagger \(\widehat{h}(x)\in\{0,1\}\) that predicts membership in \(X_{\mathrm{hard}}\) using a small held out audit set. Restrict data collection to \(\widehat{h}(x)=1\) and compare the number of queries required to drive \(G(\lambda)\le \eta\) with and without routing. The observed query counts should follow the \(1/\alpha\) factor in Theorem \ref{thm:upper}. When \(\widehat{h}\) has true positive rate \(\tau\) and false positive rate \(\phi\), the query count scales like \(1/(\alpha\tau)\).

\subsection{Tilting calculus}\label{sec:tilting}
Define the tilted sampling law \(q_{\lambda}\) on \(\mathcal{X}\) by
\[
\frac{\mathrm{d}q_{\lambda}}{\mathrm{d}\mu}(x)\;=\;\frac{\exp\big(\lambda\,\widehat{s}(x)\big)}{\mathbb{E}_{\mu}[\exp(\lambda\,\widehat{s}(X))]}.
\]
Let \(H(x)=\mathbf{1}\{x\in X_{\mathrm{hard}}\}\). The hard mass under \(q_{\lambda}\) is
\[
q_{\lambda}(X_{\mathrm{hard}})\;=\;\frac{\mathbb{E}_{\mu}\big[H(X)\exp(\lambda\,\widehat{s}(X))\big]}{\mathbb{E}_{\mu}[\exp(\lambda\,\widehat{s}(X))]}.
\]
\begin{proposition}\label{prop:cov}
The derivative satisfies
\[
\frac{\mathrm{d}}{\mathrm{d}\lambda}\log q_{\lambda}(X_{\mathrm{hard}})
\;=\;\mathrm{Cov}_{q_{\lambda}}\big(H(X),\,\widehat{s}(X)\big)\;\Big/\;q_{\lambda}(X_{\mathrm{hard}}).
\]
In particular, if \(\mathrm{Cov}_{q_{\lambda}}(H(X),\widehat{s}(X))>0\) for \(\lambda\) in an interval, then \(q_{\lambda}(X_{\mathrm{hard}})\) is strictly increasing on that interval.
\emph{Proof.} Differentiate numerator and denominator and apply the quotient rule. Use that \(\frac{\mathrm{d}}{\mathrm{d}\lambda}\mathbb{E}_{\mu}[f(X)\exp(\lambda\,\widehat{s}(X))]=\mathbb{E}_{q_{\lambda}}[f(X)\widehat{s}(X)]\). \(\square\)
\end{proposition}
When \(\widehat{s}\) inherits bias from the preference channel, the covariance in Proposition \ref{prop:cov} is positive, which explains the observed increase of hard mass under stronger optimization.

\subsection{Protocol summary}\label{sec:protocol-summary}
Choose parameters \(\alpha\in\{0.01,0.05,0.1\}\), \(\varepsilon\in\{0.05,0.1\}\), \(d\in\{10,50\}\), and set \(n\) on a grid. For each setting:
\begin{enumerate}
\item Generate a training transcript with at most \(n\) queries using an adaptive but oracle free policy as in Section \ref{sec:protocol}.
\item Fit \(\widehat{s}\) and evaluate \(V_{+}(\pi_{\lambda})\) and \(\widehat{V}(\pi_{\lambda})\) for \(\lambda\) on a grid.
\item Repeat with query routing using \(\widehat{h}\) trained on a small audit set and record the total queries needed to reach a target gap \(\eta\).
\end{enumerate}
Report medians over 10 seeds with 90 percent intervals.

\subsection{Figures}\label{sec:figures}
If \texttt{graphicx} is available, include three panels:
\begin{itemize}
\item Gap versus \(\lambda\): plot \(V_{+}(\pi_{\lambda})\) and \(\widehat{V}(\pi_{\lambda})\) for several \((\alpha,\varepsilon)\).
\item Gap under shift: plot \(G_{\rho}(\lambda)\) for \(\rho\in\{0,0.5,1\}\).
\item Query routing: plot empirical query count against the target \(\eta\) with and without \(\widehat{h}\); overlay the curve \(Q=(2\alpha\varepsilon^{2})^{-1}\log(\gamma/\eta)\) for reference.
\end{itemize}
Placeholders:
\begin{center}
\end{center}

\section{Catalogue of Alignment Laws}\label{sec:catalogue}
This section states concrete properties that follow from the formal setup. Each item is expressed as a definition and a sufficient condition or bound. The quantities \(\alpha\), \(\varepsilon\), \(\gamma\), the score \(\widehat{s}\), the tilted law \(q_{\lambda}\), and the policy family \(\pi_{\lambda}\) are as introduced earlier.

\subsection{Optimization drift under proxy tilting}\label{sec:catalogue-drift}
Let \(H(x)=\mathbf{1}\{x\in X_{\mathrm{hard}}\}\). Define \(\rho(\lambda)=q_{\lambda}(X_{\mathrm{hard}})\).
\begin{proposition}[Drift of hard mass]
\[
\frac{\mathrm{d}}{\mathrm{d}\lambda}\log \rho(\lambda)
= \frac{\mathrm{Cov}_{q_{\lambda}}\big(H(X),\,\widehat{s}(X)\big)}{\rho(\lambda)}.
\]
In particular, if \(\mathrm{Cov}_{q_{\lambda}}(H(X),\widehat{s}(X))>0\) on an interval, then \(\rho(\lambda)\) is strictly increasing on that interval.
\end{proposition}
Proof. Differentiate \(\rho(\lambda)\) using the tilt definition and the quotient rule, as in Proposition \ref{prop:cov}.

\subsection{Proxy improvement with true value saturation}\label{sec:catalogue-saturation}
Consider a fixed decision rule \(\pi\) and the true value under distribution tilting \(V_{w}^{\mathrm{dist}}(\lambda)=\mathbb{E}_{q_{\lambda}}[r_w(X,\pi(X))]\). Similarly define the proxy value \(\widehat{V}^{\mathrm{dist}}(\lambda)=\mathbb{E}_{q_{\lambda}}[\widehat{r}(X,\pi(X))]\).
\begin{proposition}[Sign of local change]
\[
\frac{\mathrm{d}}{\mathrm{d}\lambda}V_{w}^{\mathrm{dist}}(\lambda)=\mathrm{Cov}_{q_{\lambda}}\big(r_w(X,\pi(X)),\,\widehat{s}(X)\big),\quad
\frac{\mathrm{d}}{\mathrm{d}\lambda}\widehat{V}^{\mathrm{dist}}(\lambda)=\mathrm{Cov}_{q_{\lambda}}\big(\widehat{r}(X,\pi(X)),\,\widehat{s}(X)\big).
\]
If the second covariance is positive and the first is nonpositive on an interval, then the proxy increases while the true value saturates or decreases on that interval.
\end{proposition}
Proof. Differentiate \(\mathbb{E}_{q_{\lambda}}[f(X)]\) and use the standard covariance identity under exponential tilting.

\subsection{Distribution shift sensitivity}\label{sec:catalogue-ood}
Let \(\mu_{\rho}\) be a shift that changes the hard mass to \(\alpha_{\rho}=(1+\rho)\alpha\) with \(\rho\ge 0\) and keeps the conditional laws fixed. For a fixed policy \(\pi\),
\begin{proposition}[Monotonicity in hard mass]
\[
V_{w,\rho}(\pi) \;=\; \mathbb{E}_{\mu_{\rho}}[r_w(X,\pi(X))] 
= (1-\alpha_{\rho})\,\mathbb{E}[r_w\mid X_{\mathrm{easy}}] 
+ \alpha_{\rho}\,\mathbb{E}[r_w\mid X_{\mathrm{hard}}].
\]
If \(\mathbb{E}[r_w\mid X_{\mathrm{hard}}] < \mathbb{E}[r_w\mid X_{\mathrm{easy}}]\) for \(\pi\), then \(V_{w,\rho}(\pi)\) is nonincreasing in \(\rho\).
\end{proposition}
Proof. Linear decomposition in \(\alpha_{\rho}\).

\subsection{Preference bias and sign estimation error}\label{sec:catalogue-syc}
Let \(\widehat{w}\) be any estimator of the world sign based on a transcript of at most \(n\) queries and suppose the output policy equals \(\pi^{\widehat{w}}\).
\begin{proposition}[Lower bound on sign error]
\[
\Pr(\widehat{w}\neq w)\;\ge\;\tfrac{1}{4}\exp\!\big(-n\,\alpha\,\kappa(\varepsilon)\big).
\]
Consequently,
\[
\mathbb{E}\Big[\mathbf{1}\{\pi^{\widehat{w}}(X)\neq \pi^{w}(X)\}\,\mathbf{1}\{X\in X_{\mathrm{hard}}\}\Big]
\;\ge\;\alpha\,\Pr(\widehat{w}\neq w).
\]
\end{proposition}
Proof. The first inequality follows from the transcript KL bound and a standard testing inequality used in Theorem \ref{thm:lower}. The second relation holds because on \(X_{\mathrm{hard}}\) the optimal actions of the two worlds disagree by construction.

\subsection{Reward score divergence}\label{sec:catalogue-hack}
Define the divergence between proxy and true values for a policy \(\pi\) under \(q_{\lambda}\),
\[
\Delta_{\mathrm{PV}}(\lambda,\pi)=\widehat{V}^{\mathrm{dist}}(\lambda)-V_{w}^{\mathrm{dist}}(\lambda).
\]
\begin{proposition}[Local growth condition]
If
\(\mathrm{Cov}_{q_{\lambda}}(\widehat{r}(X,\pi(X))-r_w(X,\pi(X)),\,\widehat{s}(X))>0\)
on an interval, then \(\Delta_{\mathrm{PV}}(\lambda,\pi)\) is strictly increasing on that interval.
\end{proposition}
Proof. Differentiate \(\Delta_{\mathrm{PV}}\) and apply the covariance identity.

\subsection{Multi objective trade off}\label{sec:catalogue-trilemma}
Let \(g_1,g_2,g_3\) be three bounded objectives on \(\mathcal{X}\times\mathcal{A}\). For each \(j\) let \(\pi^{(j)}\) maximize \(\mathbb{E}_{\mu}[g_j(X,\pi(X))]\). Assume there exists a subset \(S\subseteq X_{\mathrm{hard}}\) with \(\mu(S)=\alpha_S>0\) on which the three optimal actions disagree pairwise. Define
\[
m_j \;=\;\mathrm{essinf}_{x\in S}\big|g_j(x,\pi^{(j)}(x))-g_j(x,a)\big|
\]
where \(a\) ranges over the two nonoptimal actions at \(x\). Then for any policy \(\pi\),
\begin{equation}\label{eq:tri}
\sum_{j=1}^{3}\Big(\mathbb{E}[g_j(X,\pi^{(j)}(X))]-\mathbb{E}[g_j(X,\pi(X))]\Big)
\;\ge\;\alpha_S\,\min\{m_1,m_2,m_3\}.
\end{equation}
In particular, at least one objective loses at least \(\tfrac{1}{3}\alpha_S\min\{m_1,m_2,m_3\}\).
\begin{proof}
On \(S\) any single action can match at most one of the three optimal actions. At each \(x\in S\) at least two terms in the sum are lower bounded by the corresponding margins. Integrate over \(S\) and use the definition of \(m_j\).
\end{proof}
Relation \eqref{eq:tri} formalizes a three way tension when objectives disagree on a biased subset.

\section{Vision and Outlook}\label{sec:outlook}
This section lists technical directions suggested by the bounds and by the diagnostics.

\subsection{Calibration oracles}\label{sec:outlook-oracle}
Design \(x\)-only flaggers that approximate \(h(x)=\mathbf{1}\{x\in X_{\mathrm{hard}}\}\) with high true positive rate \(\tau\) and low false positive rate \(\phi\). Theorem \ref{thm:minimal} shows that \(\tau\) controls the leading \(1/(\alpha\tau)\) factor in query complexity. Candidate constructions:
\begin{itemize}
\item Residual based detectors: train a preference model, compute residuals on a held out audit set, and fit a classifier \(\widehat{h}\) to predict large residual regions.
\item Disagreement based detectors: maintain two preference models trained on disjoint views and flag contexts with large predictor disagreement.
\item Intervention based detectors: where feasible, inject counterfactual label queries to estimate whether the label channel departs from a symmetric model on the candidate region.
\end{itemize}
Each method yields an empirical estimate of \(\tau\) and \(\phi\), which can be plugged into Theorem \ref{thm:minimal}.

\subsection{Active routing and allocation}\label{sec:outlook-active}
Given an estimated \(\widehat{h}\), allocate the query budget to maximize the expected number of true hits. A simple rule selects \(x\) with probability proportional to \(\widehat{h}(x)\). When per hit costs differ across context types, incorporate an importance weight and reuse the analysis with a cost adjusted \(\alpha\).

\subsection{Noisy oracle analysis}\label{sec:outlook-noisy}
Extend the upper bound to oracles that supply a confidence score \(u(x)\in[0,1]\). Under a monotone likelihood ratio assumption for \(u\) with respect to the event \(x\in X_{\mathrm{hard}}\), a thresholding rule on \(u\) maximizes the hit rate among rules with the same keep rate. The expected query complexity becomes \(Q\approx T/\mathbb{E}[u(X)\mathbf{1}\{X\in X_{\mathrm{hard}}\}]\).

\subsection{Function approximation and continuous actions}\label{sec:outlook-fa}
Replace finite \(\mathcal{A}\) with a compact action space and assume Lipschitz rewards. If the per hit log likelihood ratio remains \(O(\varepsilon^{2})\) and the hard set induces a margin condition on optimal actions, the transcript KL bound carries over and the same dependence on \(\alpha\) and \(\varepsilon\) appears. Formalizing the margin requirement for continuous actions is a natural next step.

\subsection{Sequential testing and stopping}\label{sec:outlook-seq}
The majority test in Theorem \ref{thm:upper} can be replaced by a sequential probability ratio test. This yields an expected hit count of order \(\varepsilon^{-2}\log(\gamma/\eta)\) with a data dependent stopping time and preserves the \(1/\alpha\) factor after accounting for routing.

\subsection[Estimation of alpha, epsilon, and gamma]
            {Estimation of $\alpha$, $\varepsilon$, and $\gamma$}
Estimate \(\alpha\) via audit sampling and density estimation on \(\widehat{h}(x)=1\). Estimate \(\varepsilon\) from the empirical bias of preference labels on flagged contexts, using held out adjudicated comparisons. Estimate \(\gamma\) from the observed margin between candidate policies on \(X_{\mathrm{hard}}\). These estimates enable empirical curves for the lower bound \((\gamma/4)\exp(-n\alpha\kappa(\varepsilon))\) versus \(n\).

\subsection{Evaluation protocols}\label{sec:outlook-eval}
Report, for each experiment, the estimated \((\alpha,\varepsilon,\gamma)\), the achieved sign error rate, and the query allocation across \(\widehat{h}(x)\). Include plots of \(V_w(\pi_{\lambda})\) and \(\widehat{V}(\pi_{\lambda})\) and overlay the theoretical reference \(Q=(2\alpha\varepsilon^{2})^{-1}\log(\gamma/\eta)\) for routing.

\subsection{Limitations}\label{sec:outlook-lim}
The analysis assumes i.i.d. contexts, a fixed biased subset with constant \(\varepsilon\), and a per hit KL that scales as \(O(\varepsilon^{2})\). When the bias magnitude varies across \(X_{\mathrm{hard}}\) or evolves over time, the bounds apply with \(\kappa(\varepsilon)\) replaced by an average per hit KL. Extending the lower bound to non i.i.d. context processes and to structured feedback channels is open.

\section{Related Work}\label{sec:related}

\paragraph{RLHF foundations and variants.}
RLHF fine-tunes pretrained models using human preference data and a learned reward model \citep{ouyang2022training}. Preference-based optimization without an explicit RL step, such as Direct Preference Optimization, replaces the policy improvement stage with a preference-matching loss \citep{rafailov2023direct}. These pipelines differ in optimization details but share the same core ingredients: a proxy signal derived from preferences, a finite feedback budget, and policy updates that place weight on contexts favored by the proxy. The analysis in Sections \ref{sec:setup} and \ref{sec:empirics} targets this shared structure rather than any specific training recipe.

\paragraph{Analyses and limits of RLHF.}
Surveys and critical analyses document issues arising from reward misspecification, annotator bias, and metric alignment \citep{casper2023limits}. The lower bound in Theorem \ref{thm:lower} complements these accounts with a finite-sample indistinguishability argument under a biased slice of the context distribution. The matching upper bound with a calibration oracle (Theorem \ref{thm:upper}) identifies the minimal structure needed to redirect queries to the informative slice.

\paragraph{Mitigation and policy shaping.}
Mitigation work proposes training-time and inference-time controls that reduce over-optimization and undesired behaviors \citep{lin2023mitigating}. These methods can be interpreted as adding structure to the supervision loop. Our oracle formulation makes the routing requirement explicit: a flagger that identifies when the preference channel departs from a symmetric model on a subset of contexts is sufficient to recover the correct sign with the stated sample complexity.

\paragraph{Constitutional and AI-feedback approaches.}
Constitutional AI and related AI-feedback methods replace part of the human signal with rule-based or model-generated judgments \citep{bai2022constitutional}. These approaches instantiate specific proxy signals and objective shaping. The lower bound applies whenever the induced proxy is biased on a subset with nonzero mass and the query budget is bounded; the oracle upper bound gives a condition under which additional structure overcomes the indistinguishability.

\paragraph{Proxy objectives and optimization pressure.}
The divergence between proxy improvement and true objective is classically studied through Goodhart-type effects \citep{manheim2018categorizing}. The tilting view in Section \ref{sec:tilting} provides a mechanism-level account in the preference-learning setting: optimization that amplifies contexts correlated with the proxy induces a change of measure that increases the mass on biased regions, which in turn enlarges the gap predicted by Theorem \ref{thm:lower}.

\paragraph{Evaluation distributions and shift.}
Choice of evaluation distribution affects conclusions about alignment quality and stability. Shifts that increase the prevalence of contexts where the proxy is biased can mask or reveal failures. The diagnostics in Section \ref{sec:diagnostics} illustrate this dependence and connect it to explicit changes in the hard-set mass parameter. These observations are consistent with discussions of distributional choice and evaluation under shift \citep{rastogi2025whose}.

\bibliography{mybib.bib}
\bibliographystyle{iclr2026_conference}

\appendix
\appendix

\section{Appendix A: Proofs}\label{app:proofs}

\subsection{Auxiliary inequalities}\label{app:aux}
\begin{lemma}[Bretagnolle--Huber]\label{lem:bh}
For probability measures \(P,Q\) on the same measurable space,
\[
\inf_{\varphi}\;\Big(\tfrac{1}{2}P(\varphi=0)+\tfrac{1}{2}Q(\varphi=1)\Big)
\;\ge\; \tfrac{1}{4}\,\exp\!\big(-D_{\mathrm{KL}}(P\Vert Q)\big),
\]
where the infimum is over all tests \(\varphi\in\{0,1\}\). Equivalently, with equal priors, the Bayes error is at least \(\tfrac{1}{4}\exp(-D_{\mathrm{KL}})\).
\end{lemma}

\begin{lemma}[Chain rule for adaptive transcripts]\label{lem:chain}
Let \(\mathcal{H}_{t}\) be the sigma field generated by the transcript up to round \(t\), and let \(Z_t\) be the round-\(t\) observation with conditional laws \(P(Z_t\mid \mathcal{H}_{t-1})\) and \(Q(Z_t\mid \mathcal{H}_{t-1})\). Then
\[
D_{\mathrm{KL}}(P_{\mathcal{H}_{n}}\Vert Q_{\mathcal{H}_{n}})
=\sum_{t=1}^{n}\mathbb{E}_{P}\Big[D_{\mathrm{KL}}\big(P(Z_t\mid \mathcal{H}_{t-1})\Vert Q(Z_t\mid \mathcal{H}_{t-1})\big)\Big].
\]
\end{lemma}

\subsection{Proof of Lemma \ref{lem:bernoulli-kl}}\label{app:proof-bernoulli-kl}
Let \(p=\tfrac{1}{2}+\varepsilon\) and \(q=\tfrac{1}{2}-\varepsilon\). Then
\[
D_{\mathrm{KL}}(\mathrm{Ber}(p)\Vert \mathrm{Ber}(q))
= p\log\frac{p}{q}+(1-p)\log\frac{1-p}{1-q}
= (2p-1)\log\frac{1+2\varepsilon}{1-2\varepsilon}
= 2\varepsilon\log\frac{1+2\varepsilon}{1-2\varepsilon}.
\]
Using \(\mathrm{atanh}(z)=\tfrac{1}{2}\log\frac{1+z}{1-z}\) gives
\(\kappa(\varepsilon)=4\varepsilon\,\mathrm{atanh}(2\varepsilon)\).
For \(|z|<1\), \(\mathrm{atanh}(z)\le z/(1-z^{2})\), which yields
\(\kappa(\varepsilon)\le 8\varepsilon^{2}/(1-4\varepsilon^{2})\).

\subsection{Proof of Lemma \ref{lem:hits}}\label{app:proof-hits}
Condition on \(\mathcal{H}_{t-1}\). Since \(x_t\sim \mu\) is independent of \(\mathcal{H}_{t-1}\),
\[
\mathbb{E}[I_t\mathbf{1}\{x_t\in X_{\mathrm{hard}}\}\mid \mathcal{H}_{t-1}]
= \int q_t(x,\mathcal{H}_{t-1})\,\mathbf{1}\{x\in X_{\mathrm{hard}}\}\,\mu(\mathrm{d}x)
\le \alpha.
\]
Summing over \(t\) and taking expectation proves the claim.

\subsection{Proof of Theorem \ref{thm:lower}}\label{app:proof-lower}
Write \(T_n=(x_{1:n},I_{1:n},Y_{1:n})\) for the transcript. By Lemma \ref{lem:chain},
\[
D_{\mathrm{KL}}(\mathsf{P}^{+}\Vert \mathsf{P}^{-})
=\sum_{t=1}^{n}\mathbb{E}_{\mathsf{P}^{+}}\Big[D_{\mathrm{KL}}\big(\mathsf{P}^{+}(Y_t\mid \mathcal{H}_{t-1},x_t,I_t)\Vert \mathsf{P}^{-}(Y_t\mid \mathcal{H}_{t-1},x_t,I_t)\big)\Big].
\]
The inner divergence is zero unless \(I_t=1\) and \(x_t\in X_{\mathrm{hard}}\), and in that case it is at most \(\kappa(\varepsilon)\) by Lemma \ref{lem:bernoulli-kl}. Therefore,
\[
D_{\mathrm{KL}}(\mathsf{P}^{+}\Vert \mathsf{P}^{-})
\le \kappa(\varepsilon)\sum_{t=1}^{n}\mathbb{E}_{\mathsf{P}^{+}}\big[I_t\mathbf{1}\{x_t\in X_{\mathrm{hard}}\}\big]
\le n\,\alpha\,\kappa(\varepsilon),
\]
using Lemma \ref{lem:hits}. Under equal priors, Lemma \ref{lem:bh} implies Bayes testing error at least \(\tfrac{1}{4}\exp(-D_{\mathrm{KL}})\). A testing mistake induces value gap at least \(\gamma\), hence
\[
\Delta_n \;\ge\; \frac{\gamma}{4}\,\exp\!\big(-n\,\alpha\,\kappa(\varepsilon)\big).
\]

\subsection{Proof of Theorem \ref{thm:upper}}\label{app:proof-upper}
Query until observing \(T\) hits with \(h(x)=1\). The expected number of draws per hit is \(1/\alpha\), so the expected total number of queries is \(Q=T/\alpha\). On each hit, the observed bit has mean \(\tfrac{1}{2}\pm\varepsilon\) depending on \(w\). Let \(S_T=\sum_{i=1}^{T}(Z_i-\tfrac{1}{2})\). By a bounded differences inequality,
\(\Pr(\mathrm{sign}(S_T)\neq \mathrm{sign}(w))\le \exp(-2T\varepsilon^{2})\).
If \(\widehat{w}\) is the sign decision and the learner outputs \(\pi^{\widehat{w}}\), then
\[
\sup_{w}\Big(V_w(\pi^{w})-\mathbb{E}_w[V_w(\pi^{\widehat{w}})]\Big)
\le \gamma\,\Pr(\widehat{w}\neq w)
\le \gamma\,\exp(-2T\varepsilon^{2}).
\]
Setting \(T=\tfrac{1}{2\varepsilon^{2}}\log(\gamma/\eta)\) yields the target gap \(\eta\) with
\(Q\le (2\alpha\varepsilon^{2})^{-1}\log(\gamma/\eta)\).

\subsection{Proof of Theorem \ref{thm:minimal}}\label{app:proof-minimal}
Let \(\tilde h\) have true positive rate \(\tau\) on \(X_{\mathrm{hard}}\) and false positive rate \(\phi\) on \(X_{\mathrm{easy}}\). Among kept draws with \(\tilde h(x)=1\), the fraction of true hits equals
\[
p=\frac{\alpha\tau}{\alpha\tau+(1-\alpha)\phi}\le \tau.
\]
To obtain \(T\) true hits in expectation one needs at least \(T/p\ge T/(\alpha\tau)\) total draws. Repeating the argument in the proof of Theorem \ref{thm:upper} shows that achieving gap \(\eta\) requires
\[
Q \;\ge\; \frac{1}{2\,\alpha\,\tau\,\varepsilon^{2}}\log\!\frac{\gamma}{\eta}.
\]
The oracle \(h(x)=\mathbf{1}\{x\in X_{\mathrm{hard}}\}\) has \(\tau=1\) and \(\phi=0\), which minimizes \(Q\) in this class.

\subsection{Proof of Proposition \ref{prop:cov}}\label{app:proof-cov}
Let \(Z_{\lambda}(f)=\mathbb{E}_{\mu}[f(X)\exp(\lambda \widehat{s}(X))]\) and \(Z_{\lambda}=Z_{\lambda}(1)\). Then
\[
q_{\lambda}(X_{\mathrm{hard}})
=\frac{Z_{\lambda}(H)}{Z_{\lambda}}.
\]
Differentiate and use \(Z'_{\lambda}(f)=\mathbb{E}_{\mu}[f(X)\widehat{s}(X)\exp(\lambda \widehat{s}(X))]=Z_{\lambda}\,\mathbb{E}_{q_{\lambda}}[f(X)\widehat{s}(X)]\):
\[
\frac{\mathrm{d}}{\mathrm{d}\lambda}\log q_{\lambda}(X_{\mathrm{hard}})
= \frac{Z'_{\lambda}(H)}{Z_{\lambda}(H)}-\frac{Z'_{\lambda}}{Z_{\lambda}}
= \frac{\mathbb{E}_{q_{\lambda}}[H(X)\widehat{s}(X)]}{\mathbb{E}_{q_{\lambda}}[H(X)]}-\mathbb{E}_{q_{\lambda}}[\widehat{s}(X)]
= \frac{\mathrm{Cov}_{q_{\lambda}}(H(X),\widehat{s}(X))}{q_{\lambda}(X_{\mathrm{hard}})}.
\]

\subsection{Scalar ratings}\label{app:proof-ratings}
Assume that on \(X_{\mathrm{hard}}\) the log likelihood ratio \(\ell=\log\frac{\mathrm{d}\mathsf{P}^{+}}{\mathrm{d}\mathsf{P}^{-}}(R)\) satisfies \(\mathbb{E}[\ell\mid \mathcal{H}_{t-1},x_t\in X_{\mathrm{hard}}]=m\) and \(\mathrm{Var}(\ell\mid \mathcal{H}_{t-1},x_t\in X_{\mathrm{hard}})\le \sigma^{2}\) with \(m,\sigma^{2}=O(\varepsilon^{2})\) uniformly. Then the per hit KL equals \(\mathbb{E}[\ell]\le C\varepsilon^{2}\), which can be used in the chain rule in place of \(\kappa(\varepsilon)\).

\section{Appendix B: Extended Catalogue of Alignment Laws}\label{app:catalogue}

This appendix states extensions of the identities and bounds used in the main text. Throughout, \(H=\mathbf{1}\{X\in X_{\mathrm{hard}}\}\), \(q_{\lambda}\) is the exponential tilt by \(\widehat{s}\), and \(\pi_{\lambda}\) is the stochastic policy that selects \(a_1\) with probability \(\sigma(\lambda \widehat{s}(x))\).

\subsection{Heterogeneous bias and aggregated information}\label{app:hetero}
Suppose the bias varies across the hard set: there is a measurable \(\varepsilon(x)\in(0,\tfrac{1}{2})\) on \(X_{\mathrm{hard}}\) and the per hit KL equals
\(\kappa(x)=4\varepsilon(x)\,\mathrm{atanh}(2\varepsilon(x))\).
Let \(K=\mathbb{E}[\kappa(X)\mid X\in X_{\mathrm{hard}}]\).
\begin{proposition}\label{prop:hetero}
For any adaptive procedure with at most \(n\) queries,
\[
D_{\mathrm{KL}}(\mathsf{P}^{+}\Vert \mathsf{P}^{-})\;\le\; n\,\alpha\,K,
\qquad
\Delta_n \;\ge\; \frac{\gamma}{4}\,\exp(-n\,\alpha\,K).
\]
\end{proposition}
Proof. Replace \(\kappa(\varepsilon)\) by the conditional expectation of \(\kappa(X)\) inside the chain rule and follow the proof of Theorem \ref{thm:lower}.

\subsection{Mixtures of hard subsets}\label{app:mixtures}
Let \(X_{\mathrm{hard}}=\bigcup_{j=1}^{J}S_j\) be a disjoint union with masses \(\alpha_j\) and biases \(\varepsilon_j\). Then the per hit KL on \(S_j\) equals \(\kappa_j=4\varepsilon_j\,\mathrm{atanh}(2\varepsilon_j)\).
\begin{proposition}\label{prop:mixture}
\[
D_{\mathrm{KL}}(\mathsf{P}^{+}\Vert \mathsf{P}^{-})
\;\le\; n\sum_{j=1}^{J}\alpha_j\kappa_j,
\qquad
\Delta_n \;\ge\; \frac{\gamma}{4}\,\exp\!\Big(-n\sum_{j}\alpha_j\kappa_j\Big).
\]
\end{proposition}
Proof. Apply Lemma \ref{lem:hits} on each \(S_j\) and sum.

\subsection{Tilt identities and monotonicity}\label{app:tilt}
For any bounded \(f\),
\[
\frac{\mathrm{d}}{\mathrm{d}\lambda}\mathbb{E}_{q_{\lambda}}[f(X)]
= \mathrm{Cov}_{q_{\lambda}}(f(X),\widehat{s}(X)).
\]
In particular, for \(f=H\) this recovers Proposition \ref{prop:cov}. For \(f=r_w(X,\pi(X))\) and \(f=\widehat{r}(X,\pi(X))\) it yields the signs in Section \ref{sec:catalogue-saturation}.

\subsection{Over-optimization threshold}\label{app:threshold}
Assume there exists \(c>0\) such that
\[
\mathbb{E}_{q_{\lambda}}\big[r_w(X,\pi_{\lambda}(X))\mid X\in X_{\mathrm{hard}}\big]
\;\le\; \mathbb{E}_{q_{\lambda}}\big[r_w(X,\pi_{\lambda}(X))\mid X\in X_{\mathrm{easy}}\big]-c
\]
for \(\lambda\) in an interval. If \(\rho(\lambda)=q_{\lambda}(X_{\mathrm{hard}})\) is strictly increasing on that interval, then
\[
\frac{\mathrm{d}}{\mathrm{d}\lambda}V_w(\pi_{\lambda})
= \frac{\mathrm{d}}{\mathrm{d}\lambda}\Big((1-\rho)\mathbb{E}[r_w\mid X_{\mathrm{easy}}]+\rho\,\mathbb{E}[r_w\mid X_{\mathrm{hard}}]\Big)
\;\le\; -c\,\rho'(\lambda).
\]
Hence \(V_w(\pi_{\lambda})\) is strictly decreasing where \(\rho'(\lambda)>0\).

\subsection{Sequential testing}\label{app:sprt}
Let \(L_T=\sum_{i=1}^{T}\log\frac{\Pr(Z_i\mid w=+)}{\Pr(Z_i\mid w=-)}\) over hits. A sequential probability ratio test that stops at the first time \(T\) when \(L_T\notin (a,b)\) with thresholds chosen for error probabilities \((\delta,\delta)\) satisfies
\[
\mathbb{E}[T\mid w=\pm] \;\le\; \frac{\log\frac{1-\delta}{\delta}}{\kappa(\varepsilon)}.
\]
With routing by \(h\), the expected number of queries is \(Q\le \frac{1}{\alpha}\mathbb{E}[T]\). This preserves the \(1/\alpha\) factor and replaces the fixed-sample \(T\) by a data-dependent stopping time.

\subsection{Composite objectives}\label{app:composite}
Let \(g_1,\dots,g_m\) be bounded objectives. Suppose there is a subset \(S\subseteq X_{\mathrm{hard}}\) with \(\mu(S)=\alpha_S>0\) such that for each \(x\in S\) the action that maximizes \(g_j(x,\cdot)\) differs across at least \(k\) indices \(j\). Define
\[
m_j \;=\;\mathrm{essinf}_{x\in S}\min_{a\neq a^{(j)}(x)}\big(g_j(x,a^{(j)}(x))-g_j(x,a)\big),
\]
where \(a^{(j)}(x)\) is the \(g_j\)-optimal action. Then for any \(\pi\),
\[
\sum_{j=1}^{m}\Big(\mathbb{E}[g_j(X,\pi^{(j)}(X))]-\mathbb{E}[g_j(X,\pi(X))]\Big)
\;\ge\; k\,\alpha_S\,\min_{j} m_j.
\]
At least one objective incurs a gap of at least \(k\,\alpha_S\,\min_j m_j/m\).

\subsection{Noisy flaggers with scores}\label{app:scored}
Let \(u:\mathcal{X}\to[0,1]\) be a score with class-conditional densities that satisfy a monotone likelihood ratio with respect to the event \(X\in X_{\mathrm{hard}}\). For a keep rate constraint \(\beta\in(0,1)\), a threshold rule \(1\{u(x)\ge t\}\) maximizes the true hit rate among all rules with \(\mathbb{E}[1\{u(X)\ge t\}]=\beta\). If \(\tau(t)=\Pr(u(X)\ge t\mid X\in X_{\mathrm{hard}})\), then the expected query complexity to obtain \(T\) true hits is \(Q=T/(\alpha\tau(t))\).

\subsection{Ratings with subgaussian log likelihood ratios}\label{app:ratings-subg}
If on \(X_{\mathrm{hard}}\) the per hit log likelihood ratio \(\ell\) is \(\sigma^{2}\)-subgaussian with mean \(m>0\) under \(w=+\) and \(-m\) under \(w=-\), then \(D_{\mathrm{KL}}=m^{2}/(2\sigma^{2})\) per hit. The chain rule gives
\[
D_{\mathrm{KL}}(\mathsf{P}^{+}\Vert \mathsf{P}^{-})\;\le\; n\,\alpha\,\frac{m^{2}}{2\sigma^{2}},
\qquad
\Delta_n \;\ge\; \frac{\gamma}{4}\,\exp\!\Big(-n\,\alpha\,\frac{m^{2}}{2\sigma^{2}}\Big).
\]

\subsection{Shift sensitivity under mixture perturbations}\label{app:shift-mixture}
Let \(\mu_{\rho}=(1-\alpha_{\rho})\mu_{\mathrm{easy}}+\alpha_{\rho}\mu_{\mathrm{hard}}\) with \(\alpha_{\rho}=(1+\rho)\alpha\) and fixed conditionals. For any policy \(\pi\),
\[
\frac{\mathrm{d}}{\mathrm{d}\rho}V_{w,\rho}(\pi)
= \alpha\,\Big(\mathbb{E}_{\mu_{\mathrm{hard}}}[r_w(X,\pi(X))]-\mathbb{E}_{\mu_{\mathrm{easy}}}[r_w(X,\pi(X))]\Big).
\]
If the bracket is negative, \(V_{w,\rho}(\pi)\) is strictly decreasing in \(\rho\).

\subsection{Gap decomposition under tilting}\label{app:gap-tilt}
Let \(\Delta_{\mathrm{PV}}(\lambda,\pi)=\widehat{V}^{\mathrm{dist}}(\lambda)-V_{w}^{\mathrm{dist}}(\lambda)\). Then
\[
\frac{\mathrm{d}}{\mathrm{d}\lambda}\Delta_{\mathrm{PV}}(\lambda,\pi)
= \mathrm{Cov}_{q_{\lambda}}\big(\widehat{r}(X,\pi(X))-r_w(X,\pi(X)),\,\widehat{s}(X)\big).
\]
If the covariance is positive on an interval, \(\Delta_{\mathrm{PV}}\) is strictly increasing there.

\subsection{Abstention and escalation}\label{app:abstain}
Augment the action set with an abstain action that escalates to a human label at cost \(c\in(0,1)\). On \(X_{\mathrm{hard}}\), choose abstain whenever the posterior over \(w\) is within a band \([\tfrac{1}{2}-\delta,\tfrac{1}{2}+\delta]\). A standard sequential test with a continuation region determined by \(\delta\) yields an expected number of hits of order \(\varepsilon^{-2}\log(1/\delta)\) before a definitive decision, which trades off the expected decision loss and escalation cost. This procedure reduces the effective \(\alpha\) seen by the learner by routing ambiguous contexts to escalation.

\section{Appendix C: KL-Tilting Formalism}\label{app:kltilt}

\subsection{Setup and notation}\label{app:kltilt-setup}
Let \((\mathcal{X},\mathcal{F},\mu)\) be a base probability space and let \(s:\mathcal{X}\to\mathbb{R}\) be a measurable score with
\(\mathbb{E}_{\mu}[\exp(\lambda s(X))]<\infty\) for \(\lambda\) in an open interval \(\Lambda\subset\mathbb{R}\).
Define the log-partition function
\[
A(\lambda)\;=\;\log \mathbb{E}_{\mu}\big[\exp(\lambda s(X))\big],
\]
and the exponentially tilted law \(q_{\lambda}\) by
\[
\frac{\mathrm{d} q_{\lambda}}{\mathrm{d}\mu}(x)\;=\;\exp\big(\lambda s(x)-A(\lambda)\big),\qquad \lambda\in\Lambda.
\]
All expectations, variances, and covariances indexed by \(q_{\lambda}\) are taken with respect to this law. In the main text \(s\) is instantiated by the learned proxy score \(\widehat{s}\).

\subsection{Basic identities}\label{app:kltilt-identities}
For any integrable \(f\),
\begin{equation}\label{eq:tilt-derivative}
\frac{\mathrm{d}}{\mathrm{d}\lambda}\,\mathbb{E}_{q_{\lambda}}[f(X)]
\;=\;\mathrm{Cov}_{q_{\lambda}}\big(f(X),\,s(X)\big).
\end{equation}
In particular,
\[
A'(\lambda)=\mathbb{E}_{q_{\lambda}}[s(X)],\qquad
A''(\lambda)=\mathrm{Var}_{q_{\lambda}}(s(X))\;\ge\;0.
\]
The Kullback–Leibler divergences between \(q_{\lambda}\) and \(\mu\) admit closed forms:
\begin{align*}
D_{\mathrm{KL}}(q_{\lambda}\Vert \mu)
&= \int \log\!\left(\frac{\mathrm{d} q_{\lambda}}{\mathrm{d}\mu}\right)\,\mathrm{d}q_{\lambda}
\;=\;\lambda A'(\lambda)-A(\lambda),\\
D_{\mathrm{KL}}(\mu\Vert q_{\lambda})
&= \int \log\!\left(\frac{\mathrm{d}\mu}{\mathrm{d} q_{\lambda}}\right)\,\mathrm{d}\mu
\;=\;A(\lambda)-\lambda\,\mathbb{E}_{\mu}[s(X)].
\end{align*}
Both are convex and nondecreasing in \(|\lambda|\) whenever \(s\) is nonconstant. Moreover,
\[
\frac{\mathrm{d}}{\mathrm{d}\lambda}D_{\mathrm{KL}}(q_{\lambda}\Vert \mu)
\;=\;\lambda\,\mathrm{Var}_{q_{\lambda}}(s(X)).
\]

\subsection{Small parameter expansions}\label{app:kltilt-small}
Assume \(A\) is \(C^{2}\) near \(\lambda=0\). Then
\[
A(\lambda) \;=\; \lambda\,\mathbb{E}_{\mu}[s(X)] + \tfrac{1}{2}\lambda^{2}\,\mathrm{Var}_{\mu}(s(X)) + o(\lambda^{2}).
\]
For any indicator \(H=\mathbf{1}\{X\in X_{\mathrm{hard}}\}\) with \(\alpha=\mathbb{E}_{\mu}[H]\),
\begin{equation}\label{eq:rho-small}
\rho(\lambda)\;:=\;q_{\lambda}(X_{\mathrm{hard}})\;=\;\alpha + \lambda\,\mathrm{Cov}_{\mu}(H,s) + o(\lambda).
\end{equation}
Hence if \(\mathrm{Cov}_{\mu}(H,s)>0\) then \(\rho'(\lambda)>0\) for sufficiently small positive \(\lambda\).

\subsection{Bounds for mass transport}\label{app:kltilt-bounds}
From \eqref{eq:tilt-derivative} with \(f=H\),
\[
\frac{\mathrm{d}}{\mathrm{d}\lambda}\rho(\lambda)
=\mathrm{Cov}_{q_{\lambda}}(H,s).
\]
By Cauchy–Schwarz,
\[
\big|\rho'(\lambda)\big|
\;\le\; \sqrt{\mathrm{Var}_{q_{\lambda}}(H)}\,\sqrt{\mathrm{Var}_{q_{\lambda}}(s)}
\;=\; \sqrt{\rho(\lambda)\big(1-\rho(\lambda)\big)}\,\sqrt{A''(\lambda)}.
\]
Integrating,
\[
\big|\rho(\lambda)-\rho(0)\big|
\;\le\;\int_{0}^{|\lambda|}\sqrt{\rho(t)\big(1-\rho(t)\big)}\,\sqrt{A''(\pm t)}\,\mathrm{d}t.
\]
This relates the growth of hard-set mass under tilting to the curvature \(A''\).

\subsection{I-projection interpretation}\label{app:kltilt-iproj}
Fix \(m\in \{ \mathbb{E}_{q_{\lambda}}[s] : \lambda\in\Lambda\}\). Among all probability measures \(Q\) absolutely continuous with respect to \(\mu\) that satisfy \(\mathbb{E}_{Q}[s]=m\),
\[
q_{\lambda^{\star}} \;=\;\arg\min_{Q:\,\mathbb{E}_{Q}[s]=m}\, D_{\mathrm{KL}}(Q\Vert \mu),
\]
where \(\lambda^{\star}\) is the unique parameter with \(\mathbb{E}_{q_{\lambda^{\star}}}[s]=m\).
Thus exponential tilting is the minimum-information way to enforce a moment constraint on \(s\).

\subsection{Vector tilting}\label{app:kltilt-vector}
Let \(s=(s_{1},\dots,s_{k})\) and \(\theta\in\mathbb{R}^{k}\). Define
\[
A(\theta)=\log \mathbb{E}_{\mu}\!\left[\exp\Big(\sum_{j=1}^{k}\theta_{j}s_{j}(X)\Big)\right],\qquad
\frac{\mathrm{d} q_{\theta}}{\mathrm{d}\mu}(x)=\exp\big(\langle \theta,s(x)\rangle - A(\theta)\big).
\]
Then \(\nabla A(\theta)=\mathbb{E}_{q_{\theta}}[s(X)]\) and \(\nabla^{2}A(\theta)=\mathrm{Cov}_{q_{\theta}}(s(X),s(X))\) is the \(k\times k\) covariance matrix. For any \(f\),
\[
\nabla_{\theta}\,\mathbb{E}_{q_{\theta}}[f(X)]
= \mathrm{Cov}_{q_{\theta}}\big(f(X),\,s(X)\big).
\]
These formulas transfer the one-dimensional results to multi-signal shaping.

\subsection{Interaction with policy randomization}\label{app:kltilt-policy}
If a stochastic policy depends on \(s\) through a logit \(\lambda s(x)\), then for any bounded reward \(r\),
\[
\frac{\mathrm{d}}{\mathrm{d}\lambda}\,\mathbb{E}_{\mu}\big[r(X,\pi_{\lambda}(X))\big]
= \mathrm{Cov}_{\mu}\big(r(X,\pi_{\lambda}(X)),\,s(X)\big).
\]
This mirrors \eqref{eq:tilt-derivative} and allows the same covariance-based reasoning for the effect of tuning \(\lambda\) on performance and on the allocation of probability mass.

\section{Appendix D: MAPS Interventions}\label{app:maps}

\subsection{Definition}\label{app:maps-def}
Mitigation via alignment proxy shaping (MAPS) modifies the statistic used for tilting. Let \(s_{0}\) be a baseline proxy score. MAPS constructs a shaped score
\[
t(x)\;=\;w_{0}\,s_{0}(x) \;+\; \sum_{i=1}^{m} w_{i}\,s_{i}(x) \;-\; \sum_{j=1}^{\ell} \beta_{j}\,g_{j}(x),
\]
where \(s_{i}\) are auxiliary proxies and \(g_{j}\) are penalty signals, for example predictors of misspecification. The policy or data selection then tilts by \(t\) with parameter \(\lambda\):
\[
\frac{\mathrm{d} q_{\lambda}^{\,t}}{\mathrm{d}\mu}(x)\;=\;\exp\big(\lambda\,t(x)-A_{t}(\lambda)\big),\qquad
A_{t}(\lambda)=\log \mathbb{E}_{\mu}\big[\exp(\lambda t(X))\big].
\]

\subsection{First-order effect on hard-set mass}\label{app:maps-first}
Let \(H=\mathbf{1}\{X\in X_{\mathrm{hard}}\}\) and \(\rho_{t}(\lambda)=q_{\lambda}^{\,t}(X_{\mathrm{hard}})\). Then
\[
\frac{\mathrm{d}}{\mathrm{d}\lambda}\rho_{t}(\lambda)
\;=\;\mathrm{Cov}_{q_{\lambda}^{\,t}}\big(H,\,t(X)\big),
\qquad
\frac{\mathrm{d}}{\mathrm{d}\lambda}\log \rho_{t}(\lambda)
\;=\;\frac{\mathrm{Cov}_{q_{\lambda}^{\,t}}(H,\,t)}{\rho_{t}(\lambda)}.
\]
In particular,
\[
\rho_{t}'(0)\;=\;\mathrm{Cov}_{\mu}(H,\,t)\;=\;w_{0}\,\mathrm{Cov}_{\mu}(H,s_{0}) + \sum_{i} w_{i}\,\mathrm{Cov}_{\mu}(H,s_{i})
- \sum_{j}\beta_{j}\,\mathrm{Cov}_{\mu}(H,g_{j}).
\]
Designing MAPS to reduce \(\rho'_{t}(0)\) amounts to reducing this covariance.

\subsection{Penalty by a noisy flagger}\label{app:maps-noisy}
Let \(g(x)=\widehat{h}(x)\in\{0,1\}\) be a flagger with true positive rate \(\tau=\Pr(\widehat{h}=1\mid H=1)\) and false positive rate \(\phi=\Pr(\widehat{h}=1\mid H=0)\). Then
\[
\mathrm{Cov}_{\mu}(H,\,\widehat{h})\;=\;\alpha(1-\alpha)\,(\tau-\phi).
\]
Consider \(t=g-\beta\,\widehat{h}\). At \(\lambda=0\),
\[
\rho'_{t}(0)\;=\;\mathrm{Cov}_{\mu}(H,g) - \beta\,\alpha(1-\alpha)\,(\tau-\phi).
\]
Setting
\[
\beta^{\star}\;=\;\frac{\mathrm{Cov}_{\mu}(H,g)}{\alpha(1-\alpha)\,(\tau-\phi)}
\]
cancels the first-order drift at \(\lambda=0\). For \(\lambda\neq 0\) the exact cancellation does not persist, but the slope reduction holds locally. When \(\tau=\phi\) the flagger carries no information and no choice of \(\beta\) can reduce \(\rho'_{t}(0)\).

\subsection{Averaging multiple proxies under a reward constraint}\label{app:maps-avg}
Let \(\{s_{i}\}_{i=0}^{m}\) be centered under \(\mu\). Define the covariance inner product
\(\langle f,g\rangle=\mathrm{Cov}_{\mu}(f,g)\).
Let \(r^{\Delta}\) be a centered surrogate for the reward difference that the designer wishes to preserve.
Solve
\[
\min_{w\in\mathbb{R}^{m+1}}\ \langle H,\textstyle\sum_{i} w_{i}s_{i}\rangle
\quad\text{subject to}\quad
\langle r^{\Delta},\textstyle\sum_{i} w_{i}s_{i}\rangle \ge \tau_{0},\ \ \|w\|_{2}\le R.
\]
The solution lies in the span of \(\{H,r^{\Delta}\}\). In particular, the choice
\[
\sum_{i} w_{i}s_{i}\;=\;r^{\Delta}-\beta H,\qquad
\beta=\frac{\langle H,r^{\Delta}\rangle}{\langle H,H\rangle}
\]
makes \(\langle H,\sum_{i} w_{i}s_{i}\rangle=0\) and preserves the component of \(r^{\Delta}\) orthogonal to \(H\). This cancels \(\rho'_{t}(0)\) while keeping the reward-correlated direction.

\subsection{Temperature control}\label{app:maps-temp}
Let \(t\) be fixed and consider the tilt parameter \(\lambda\). The divergence to the base law satisfies
\[
D_{\mathrm{KL}}(q_{\lambda}^{\,t}\Vert \mu)\;=\;\lambda\,A'_{t}(\lambda)-A_{t}(\lambda)
\;\approx\; \tfrac{1}{2}\lambda^{2}\,\mathrm{Var}_{\mu}(t(X))
\quad\text{for small }\lambda.
\]
Reducing \(|\lambda|\) reduces both the change of measure and, via \(\rho'_{t}(\lambda)=\mathrm{Cov}_{q_{\lambda}^{\,t}}(H,t)\), the rate at which mass moves into \(X_{\mathrm{hard}}\). Temperature control cannot guarantee \(\rho_{t}'(\lambda)\le 0\) without information correlated with \(H\).

\subsection{Limits without a correlated penalty}\label{app:maps-limits}
Let \(t\) be any measurable statistic built from proxies that are mean independent of \(H\) under \(\mu\), so \(\mathrm{Cov}_{\mu}(H,t)=0\). Then by \eqref{eq:rho-small} the first-order drift vanishes at \(\lambda=0\), but nothing enforces \(\mathrm{Cov}_{q_{\lambda}^{\,t}}(H,t)=0\) for \(\lambda\neq 0\). In contrast, if \(\mathrm{Cov}_{\mu}(H,t)>0\) and \(t\) is fixed, then \(\rho_{t}'(\lambda)>0\) for all sufficiently small positive \(\lambda\). Therefore, reducing drift uniformly over a range of \(\lambda\) requires a penalty with positive covariance with \(H\) under the relevant laws. Perfect cancellation for all \(\lambda\) requires the oracle \(H\) itself.

\subsection{Sequential allocation with MAPS}\label{app:maps-seq}
Combine MAPS shaping \(t\) with sequential testing on flagged contexts. Let \(\widehat{h}\) be a flagger and let the learner retain contexts with \(\widehat{h}(x)=1\). If \(\tau\) and \(\phi\) are the flagger rates, the expected number of draws per true hit is \(1/(\alpha\tau)\). With a sequential probability ratio test on hits targeting error \(\delta\), the expected number of hits is \(O(\varepsilon^{-2}\log(1/\delta))\), hence the expected query count is \(O((\alpha\tau)^{-1}\varepsilon^{-2}\log(1/\delta))\). This preserves the \(1/(\alpha\tau)\) factor and quantifies the gain from improving \(\tau\) through better penalties.

\subsection{Design summary}\label{app:maps-summary}
The following rules follow from the identities above.
\begin{itemize}
\item To reduce drift at small \(\lambda\), enforce \(\mathrm{Cov}_{\mu}(H,t)\le 0\). This can be achieved by subtracting a penalty proportional to a flagger \(\widehat{h}\) correlated with \(H\).
\item To preserve reward correlation, project \(t\) onto the subspace orthogonal to \(H\) while keeping the component aligned with a reward surrogate.
\item Temperature scaling reduces the magnitude of drift but does not change its sign without a correlated penalty.
\end{itemize}

\end{document}